\pdfoutput=1

\documentclass[11pt]{article}

\usepackage[]{acl}

\usepackage{times}
\usepackage{latexsym}
\usepackage{float}

\usepackage[T1]{fontenc}

\usepackage[utf8]{inputenc}

\usepackage{microtype}

\usepackage{graphicx}
\usepackage{booktabs}
\usepackage{mathtools}
\newcommand{\ex}[1]{``#1''}
\newcommand{\arc}[1]{\textsc{{#1}}}
\newcommand{\fm}[1]{\textit{#1}}

\usepackage{colortbl}
\usepackage{tikz}
\usetikzlibrary{shapes,arrows}
\usepackage{pdflscape}
\usepackage{geometry}
\usetikzlibrary{decorations.text}
\usepackage{xcolor}
\usepackage{tikz-dependency}

\usepackage{todonotes}
\makeatletter
\newcommand*\iftodonotes{\if@todonotes@disabled\expandafter\@secondoftwo\else\expandafter\@firstoftwo\fi}
\makeatother

\title{Compositional Generalization in Dependency Parsing}

\author{Emily Goodwin\thanks{~~Majority of work completed during internship at Element AI, now ServiceNow Research. Corresponding author, emily.goodwin@mail.mcgill.ca} \textsuperscript{1,3}
  Siva Reddy\textsuperscript{1,2,3,6}
  Timothy J. O'Donnell\textsuperscript{1,3,5} 
  Dzmitry Bahdanau\textsuperscript{4, 5}\\ 
  \textsuperscript{1}Department of Linguistics
  \textsuperscript{2}School of Computer Science~~~McGill University,
  Canada \\
     \textsuperscript{3}Quebec Artificial Intelligence Institute (Mila) 
     ~~~\textsuperscript{4}ServiceNow Research \\
     ~~~~\textsuperscript{5}Canada CIFAR AI Chair ~~~~ \textsuperscript{6}Facebook CIFAR AI Chair \\
}

\begin{document}
\maketitle
\begin{abstract}
  Compositionality\textemdash the ability to combine familiar units like words
  into novel phrases and sentences\textemdash has been the focus of intense
  interest in artificial intelligence in recent years.  To test
  compositional generalization in semantic parsing,
  \citet{keysers2020measuring} introduced Compositional Freebase
  Queries (CFQ). This dataset maximizes the similarity between the
  test and train distributions over primitive units, like words, while
  maximizing the \textit{compound divergence}---the dissimilarity
  between test and train distributions over larger structures, like
  phrases. Dependency parsing, however, lacks a compositional generalization
  benchmark. In this work, we introduce
  a gold-standard set of dependency parses for CFQ, and use this to
  analyze the behavior of a state-of-the art dependency parser
  \citep{qi2020stanza} on the CFQ dataset. We find that increasing
  compound divergence degrades dependency parsing performance,
  although not as dramatically as semantic parsing performance.
  Additionally, we find the performance of the dependency parser does
  not uniformly degrade relative to compound divergence, and the
  parser performs differently on different splits with the same
  compound divergence.  We explore a number of hypotheses for what
  causes the non-uniform degradation in dependency parsing
  performance, and identify a number of syntactic structures that
  drive the dependency parser's lower performance on the most
  challenging splits.
\end{abstract}

\section{Introduction}
    \begin{figure}[t]  
\centering  

\includegraphics[width = \linewidth, keepaspectratio]{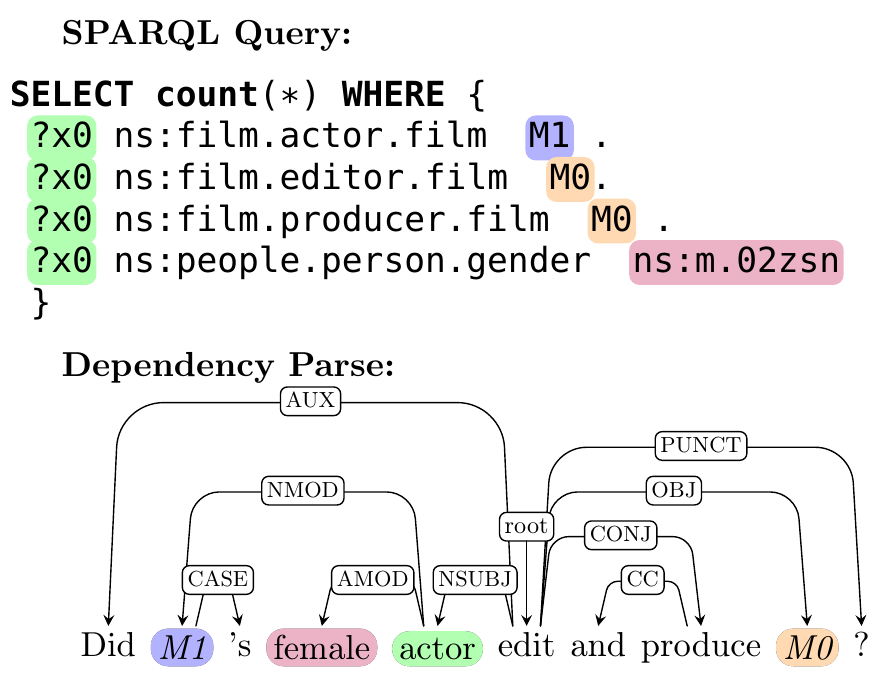} 
 
 \caption{An example question from the CFQ dataset, with the associated SPARQL query and dependency parse.}  
\label{fig:fptrhc}  
\end{figure}  
\label{sec:intro}
People understand novel combinations of familiar words in part due to
the principle of \fm{compositionality}: We expect the
meaning of a phrase to be a predictable composition of the meanings of
its parts. Unlike humans, many neural models fail to generalize 
compositionally; a
growing interest in this area has led to novel architectures and
datasets designed to test compositional generalization (see \textsection~\ref{sec:relatedwork}).

One recently-introduced semantic parsing dataset, Compositional
Freebase Queries (CFQ), consists of English questions with
corresponding database queries written in SPARQL. Figure~\ref{fig:fptrhc} 
shows an example question and SPARQL query. To test
compositional generalization, CFQ includes test and train sets with a
highly similar distribution of primitive units (like words) and
increasingly divergent distribution of larger \fm{compound} units
(like phrases). The most challenging of these splits, with the highest
compound divergence, are dubbed \fm{maximum compound divergence} (MCD)
splits. 

Although CFQ has proven to be a valuable resource, the difficulty of the splits 
appears to be influenced by factors other than compositional generalization.
First, some evidence suggests that the complexity of the SPARQL output 
is in part responsible for CFQ performance \citep{furrer2020compositional, herzig_unlocking_2021}.  
Furthermore, splits of the same compound divergence are not equally difficult. 
One possible explanation is a difference in the syntactic constructions of different splits; however, this has not yet been explored in CFQ.
To address these issues, we created a dependency-parsing version of CFQ.
Using our dataset, we evaluated a state-of-the-art dependency parser for
compositional generalization, and used the dependency annotations to identify syntactic
structures predictive of parsing failure on each MCD split.

We found that the dependency parser is more robust to increased
compound divergence than the semantic parser, but performance still
decreased with higher compound divergence. 
We also found the dependency parser, like semantic parsers, varied 
widely in performance on different splits of the same compound divergence. 
Finally, we found that a small number (less than seven)
of syntactic constructions seem to drive the difficulty of the MCD splits. 
Our dataset is publically available on github.\footnote{\href{https://github.com/emilygoodwin/CFQ-dependencies}{https://github.com/emilygoodwin/CFQ-dependencies}}

\subsection{Motivation for Dependency Parsing}
\label{sec:motivation}
In this section, we discuss three problems of CFQ, and our motivation
for studying compositional generalization in dependency parsing. 

First, CFQ is hard: seq2seq models trained from scratch score at most 12\% on MCD2 and
MCD3 sets \cite{CFQleaderboard}. Because of its difficulty, CFQ may lack sensitivity to
capture small but significant progress in neural modelling of
compositionality.  Second, recent work shows that CFQ's difficulty is
in part due to the output representation being raw SPARQL: Models
perform better when outputs are replaced with compressed versions of
SPARQL, that are more aligned with the natural-language-like questions
\citep{furrer2020compositional, herzig_unlocking_2021}. In interpreting
performance on CFQ, we might be conflating challenges of compositional generalization with 
challenges related to the output representation. 

Third, different splits of the same compound divergence vary widely in difficulty: seven 
of the nine semantic parsers currently listed on the leaderboard 
perform at least twice as well on MCD1 as MCD 2, despite the splits having the same compound
divergence \cite{CFQleaderboard}.
Performance on CFQ is thus heavily influenced by some factor
about the splits other than compound divergence.
  
Finally, CFQ lacks a description of the specific syntactic generalizations 
tested by each split. Related benchmarks, like COGS \citep{kim2020cogs} and
CLOSURE \citep{bahdanau2020closure}, test a clearly-defined set of generalizations
(for example, training a noun in subject position and testing in object position). 
CFQ splits, by contrast, optimize a gross metric over the distribution of all 
syntactic compounds in the dataset. This complicates in-depth analyses of CFQ 
results: For a particular split, it is unclear what syntactic constructions 
are tested in out-of-distribution contexts. Meanwhile, for a particular test 
sentence, it is unclear which of its syntactic structures caused the model to fail.

To address the issues with the CFQ semantic parsing benchmark,
 we studied compositional generalization in syntactic parsing. 
While syntactic parsing is simpler than mapping to a complete meaning 
representation, a language-to-SPARQL semantic parser must understand the
question's syntax. For example, to generate the triple \texttt{?x0
  ns:film.editor.film M0} in the SPARQL query shown in Figure~\ref{fig:fptrhc}, a semantic parser must first identify that ``actor''
is the subject of ``edit''.

We chose dependency trees as the target syntactic formalism due to
the maturity of the universal dependencies annotation standard, the
popularity of dependency trees among the NLP practitioners, and the
availability of popular high-performance software such as Stanza
\cite{qi2020stanza}.
Importantly, dependency parsing does not require auto regressive
models; instead, graph dependency parsers independently predict edge
labels. This different way of employing deep learning for parsing
has the additional advantage of allowing us to separate the challenge of compositional generalization 
from challenges related to auto regressive
models' teacher forcing training. Finally, having gold dependency
annotations for CFQ questions enables detailed analysis of the
relation between the model errors and syntactic discrepancies that are
featured by the MCD splits.

\section{Compound Divergence in CFQ}
\label{subsec:mcd}
CFQ is designed to test compositional generalization by combining familiar 
units in novel ways. To ensure the primitive units are familiar to the learner, CFQ test
and train sets are sampled in a way that ensures a low divergence in
the frequency distribution of \fm{atoms}. Here, atoms refers to
individual predicates or entities, (like \ex{produced} or
\ex{Christopher Nolan}), and the rules used to generate questions.

To ensure the compounds in test are novel, train and test sets were sampled
in a way that ensures higher divergence between the frequency
distribution of compounds, weighted to prevent
double-counting of any nested compounds which co-occur frequently.

\citet{keysers2020measuring} released dataset splits with compound 
divergence on a scale between 0 (a random split) and .7 (Maximum Compound Divergence, or MCD, splits).

\section {Corpus Construction: Dependency Parses for CFQ}
\label{subsec:goldparses}
To train a dependency parser and analyze syntactic structures in the
CFQ dataset, we created a corpus of gold dependency parses. Because the
questions in CFQ are synthetically generated, we were able to write a 
full-coverage context-free grammar for the
CFQ language (see Appendix~\ref{sec:CFG}). Using this grammar, and the
chart parser available in Python's natural language toolkit, we
generated a constituency parse for each question. Finally, we designed an
algorithm to map to the dependency parse.

To map from constituency to dependency parses, we wrote a
dependency-mapping rule for each production rule in the CFG
\citep{collins2003head}. Each dependency rule describes the
dependency relation between the elements in the constituent; for
example, if the production rule is
$\mathrm{VP \longrightarrow V\textit{ }NP}$, the dependency-mapping
rule connects the head of the right-hand node (the head of NP) as a
dependent of the left-hand node (the V), with the arc label
\arc{obj}. We follow version two of the Universal Dependencies Corpus
annotation standards
\citep{nivre-etal-2020-universal},\footnote{\href{www.universaldependencies.org}{www.universaldependencies.org}}
but simplify the categorization of nominal subjects for active and
passive verbs into one category (\arc{nsubj}), and do not include part
of speech tags in the dataset.

Our algorithm then recursively walks the constituency
tree from bottom to top, mapping non-head children of each node to
their syntactic heads and passing the head of each constituent up the
tree.  A number of sentences in the CFQ dataset exhibit dependency
structures which cannot be directly read off the constituency parse in
this manner: Such \fm{right-node-raising} constructions involve a word
without a syntactic head in the immediate constituent. For example, in
\ex{Was \textit{Tonny} written by and executive produced by Mark
  Marabella?}  the first instance of \ex{by} is a dependent of
\ex{Mark Marabella}, but its immediate constituent is \ex{directed
  by}. To handle right-node raising cases, our
dependency-mapping algorithm identifies prepositions with no head in
the immediate constituent, and passes them up the tree until they can be
attached to their appropriate syntactic head.

Finally, we performed a form of anonymization on the questions,
replacing entities with single-word proper names. This reflects the
anonymization strategy used in \citet{keysers2020measuring}, 
and prevents the dependency parser from
failing because of named entities with particularly complex internal
syntax (for example, \ex{Did a Swedish film producer edit
  \textit{Giliap} and \textit{Who Saw Him Die?}})

The experiments in this paper are based on the original CFQ
splits. However, these validation sets 
are constructed from the same distribution as the test sets;
some information about the test distribution is therefore available
during train. To ensure that the model only had access to the training
distribution during the training phase, we followed the suggestion of
\citet{keysers2020measuring} and discarded the MCD
validation sets,\footnote{We also
  re-sampled a validation set from the random split training set, so
  that MCD and random splits are trained on the same amount of data.} randomly sampling $20 \%$ of the training data to use
instead (see \textsection~5.1 of that paper for more details). The
resulting splits have $11,968$ test sentences and $76,595$ train
sentences.

\section{Compound Divergence Effect on Dependency Parsing}
\subsection{Training Stanza}
\label{subsec:trainingStanza}
To evaluate the effect of compound divergence on dependency parsing,
we used Stanza \citep{qi2020stanza}, a state-of-the-art dependency
parser, on the gold label dependency parses described in
\textsection\ref{subsec:goldparses}. We trained Stanza five times on
each of 22 splits from the CFQ release: one random split (which has a
compound divergence of 0), 18 splits
with increasing compound divergence (ranging from .1 to .6) and three
MCD splits (divergence of .7).

To evaluate performance on each test set, we used the CoNLL18 shared
task dependency parsing challenge evaluation script \cite{UDevaluation}, which
gives a Labeled Attachment Score (LAS) and Content-word Labeled
Attachment Score (CLAS), reflecting how many of the total dependency
arcs in the test set were correctly labeled, and how many of the arcs
connecting content words were correctly labeled, respectively.

In addition, we calculated the percentage of test questions for which
every content word arc was correctly labeled, which we call \fm{Whole
  Sentence Content-word Labeled Attachment Score} (WSCLAS). This
all-or-nothing evaluation scheme for each sentence more closely
resembles the exact-match accuracy of semantic parser evaluation.\footnote{The code to calculate WSCLAS is also available at \href{https://github.com/emilygoodwin/CFQ-dependencies}{https://github.com/emilygoodwin/CFQ-dependencies}}

\subsection{Dependency Parsing Results}
\label{sec:allSplitsresults}

\begin{figure}[ht]
    \includegraphics[width = .45\textwidth, keepaspectratio]{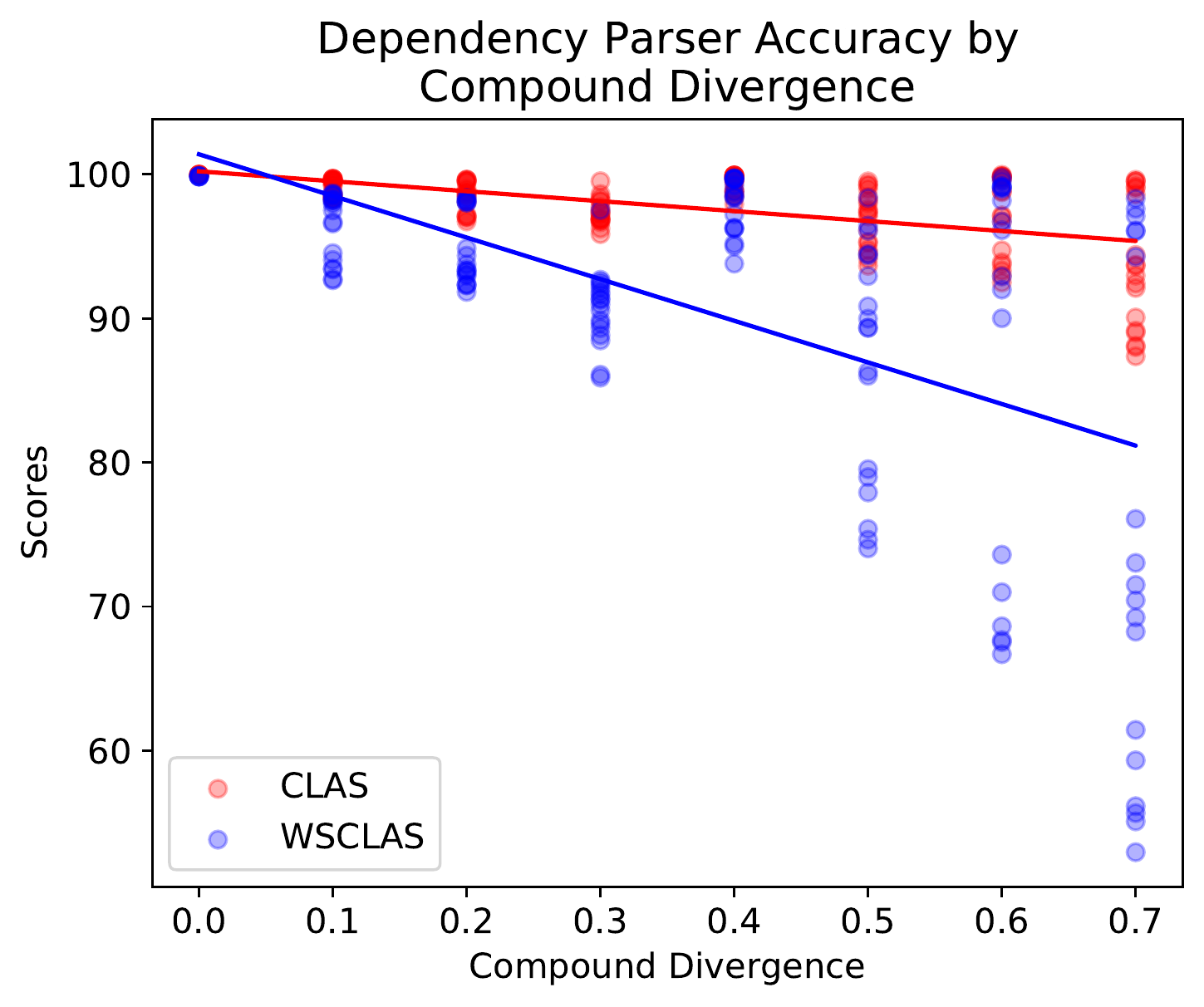}
    \caption{The effect of compound divergence on Content-word Labeled Attachment Score (CLAS) and Whole-Sentence Content Labeled Attachment Score (WSCLAS).}
    \label{fig:scatterplot}
\end{figure}

We plot Stanza's performance as a function of the split compound 
divergence in Figure~\ref{fig:scatterplot}. 
Increasing compound divergence had a negative effect on performance: 
Stanza's accuracy on the random split (zero compound divergence) was near perfect, with an average CLAS of $99.98\%$ and WSCLAS of $99.89\%$. Meanwhile, accuracy on the three MCD splits (divergence of .7) dropped to an average CLAS of $92.85\%$ and WSCLAS of $74.92\%$. 

A linear regression predicting CLAS found a slope of $-6.91$, and
predicting WSCLAS found a slope of $-28.89$; in other words, for each
.1 increase in compound divergence the linear model predicts a
$2.889\%$ lower WSCALS, and $.691\%$ lower CLAS. These linear models
are also shown in Figure~\ref{fig:scatterplot}.

\begin{table}[ht]
\centering
\resizebox{\linewidth}{!}{
\begin{tabular}{c c c c c}
\toprule
& \multicolumn{3}{c}{Dependency Parser} & Semantic Parser\\
Split & WSCLAS & CLAS & LAS & Exact Match\\
& \small{mean (sd)} & \small{mean (sd)} & \small{mean (sd)} & \small{mean ($95\%$ conf interval)}\\
\midrule
MCD1 & 96.57 $\pm{1.31}$ & 99.38 $\pm{0.29}$ & 99.64 $\pm{0.16}$& 37.4 $\pm{2.2}$\\
\rowcolor{gray!6} MCD2 & 71.42 $\pm{2.59}$& 91.53  $\pm{1.00}$ & 93.28 $\pm{0.88}$& 8.1$\pm{1.6}$\\
MCD3 & 56.76 $\pm{2.81}$ & 87.66  $\pm{0.93}$ & 90.87 $\pm{1.01}$&11.3 $\pm{0.3}$ \\
\rowcolor{gray!6} Random & 99.89 $\pm{0.01}$ & 99.98$\pm{0.00}$ & 99.99 $\pm{0.00}$& 98.0 $\pm{0.3}$\\
\bottomrule
\end{tabular}}
\caption{Stanza's performance on MCD splits in terms of Whole Sentence Content-word Labeled Attachment Score (WSCLAS), Content-word Labeled Attachment Score (CLAS), and Labeled Attachment Score (LAS). Means and standard deviations are calculated over five randomly-seeded runs. The semantic parsing scores are reproduced from \citep{keysers2020measuring}; the mean exact-match of 5 experiments with $95\%$ confidence intervals is reported for each MCD split in their github repository.\footnote{https://github.com/google-research/google-research/tree/master/cfq}
\label{tbl:mcd_scores}}
\end{table}

We note, however, two exceptions to the
generally negative relationship between compound divergence and
accuracy, which indicate that other characteristics of the test set
have a large effect on accuracy. First, all splits with a target
compound divergence of $.4$ performed stronger than those at divergence
$.3$ and $.2$. Secondly, we observed considerable variation in performance
on different splits that have the same compound divergence,
particularly the MCD splits.

Stanza's performance on the three maximum-compound-divergence splits
and one random split is shown in Table~\ref{tbl:mcd_scores}.
While all three MCD splits were harder than the random split, performance
varied from $96.57\%$ WSCLAS (MCD1) to $56.76\%$ WSCLAS (MCD3). Thus,
while compound divergence is a factor in performance, idiosyncrasies
in the individual splits also have large effects on performance.

Finally, we note that while  Stanza was more robust to compound divergence 
than the semantic parser, it also ranked the splits differently in
difficulty. Table~\ref{tbl:mcd_scores} reproduces mean accuracies from \citet{keysers2020measuring}'s strongest-performing semantic parser, a universal transformer \citep{dehghani2019universal}. The universal transformer's exact-match is lower than Stanza's WSCLAS on every MCD split. Additionally, while Stanza performed worst on MCD3, the universal transformer and most other semantic parsers in the CFQ leaderboard performed worst on MCD2 \cite{CFQleaderboard}.
In the next sections, we explore what causes the variation in
performance on different MCD splits.

\section{Construction Complexity and the MCD Splits}
The compound divergence metric treats all compounds of any number of
words identically; therefore, the differences between the MCD
splits may be driven by differing distributions of compounds of
different complexities. In this section, we show that this is not the case. 
We first describe how we characterize syntactic constructions
using the dependency annotations.

\subsection{Syntactic Constructions}
\label{sec:constructions}
\begin{figure}[ht]
    \includegraphics[width = .5\textwidth, keepaspectratio]{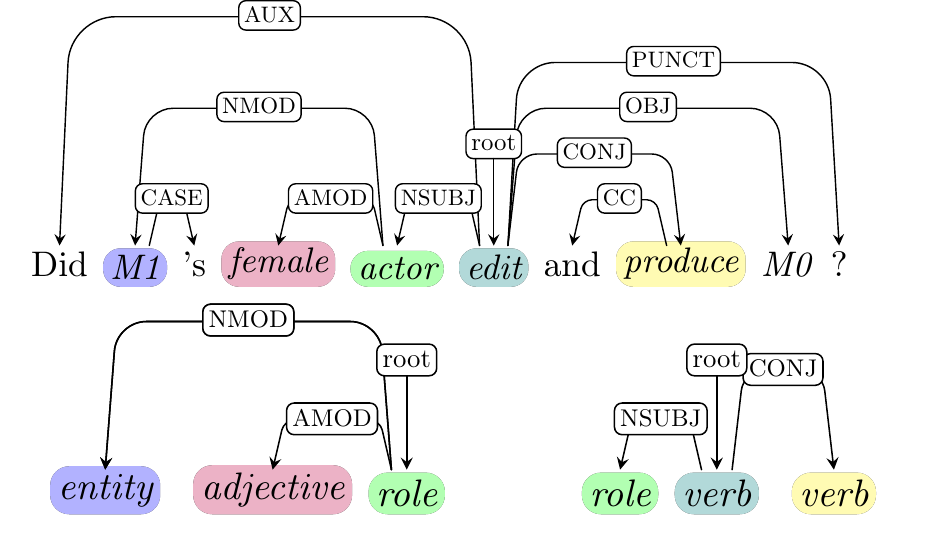}
    \caption{A dependency parse and two of its subtrees}
    \label{fig:subtrees}
\end{figure}

We explored differences in the distributions of syntactic constructions
by looking at a restricted set of the subtrees of each dependency
parse, which we will now describe. 

  With respect to any \fm{target node} in
the corpus, we consider a syntactic construction to be any subtree
that consists of that target node together with a
\fm{constituent-contiguous} subset of the target node's immediate
children. Here, constituent-contiguous means the subsets of child nodes
which are heads of phrases that are adjacent to one another or to the
target node in the string. We include only the immediate children in
the subtree (excluding their descendants). 
We also replace words with their category label in CFQ: in addition to traditional
parts of speech like \fm{verb} and \fm{adjective}, the category labels
include nominal categories \fm{role} (which occurs in possessive
constructions like \ex{mother} in \ex{Alice's mother}), \fm{entity}
for proper nouns, and \fm{noun} for common nouns.

For the analyses in this and the following section, we extract every syntactic construction for every dependency parse in our corpus, and compare their \fm{complexity}. 
We define complexity to be the number of arcs in the subtree, discounting the dummy \arc{root} arc. 
Two of the subtrees for sentence
\ex{Did \textit{M1}'s female actor edit and produce \textit{M0}?} are
shown in Figure~\ref{fig:subtrees} (these subtrees have a complexity
of two). Table~\ref{tbl:dataset_statistics} shows the number of unique
constructions in each test and train set.\begin{table}[ht]
\centering
\resizebox{\linewidth}{!}{
\begin{tabular}{c c c c c}
\toprule
& \multicolumn{2}{c}{Total Sentences} & \multicolumn{2}{c}{Unique Constructions} \\
 & Train & Test & Train & Test\\
\midrule
MCD1  & 76,595 & 11,968 & 2,093 & 2,048 \\
\rowcolor{gray!6} MCD2  & 76,595 & 11,967 & 2,006 & 1,884 \\
MCD3 & 76,595 & 11,968 & 2,300 & 1,823 \\
\rowcolor{gray!6} Random & 76,596 & 11,967 & 4,082 & 3,251\\
\bottomrule
\end{tabular}}
\caption{Number of sentences and unique constructions for each test and train set in our experiments (see \textsection~\ref{sec:constructions} for an explanation of constructions). 
\label{tbl:dataset_statistics}}
\end{table}

\subsection{Analysis of MCD Splits}
\label{sec:split_tree_distributions}
One possible source of the differences between MCD splits may be that
they differ in their distributions of subtrees at differing
complexities.
In this section, we present two analyses
showing that this is not the case.

In our first analysis, we analyzed the distance between test and train
distribution for each split. To do this we calculated the
Jensen-Shannon (JS) distance between the test and train histograms of
syntactic constructions at differing complexities.\footnote{The JS
  distance for histograms $p$ and $q$ is defined as
\begin{equation}
\sqrt{\frac{D(\;p\|m\;) + D(\;q\|m\;)}{2}}
\end{equation}
where $m$ is the pointwise mean of $p$ and $q$, and $D$ is the
Kullback-Leibler divergence.}

\begin{figure}[ht]
    \centering
    \includegraphics[width = .45\textwidth]{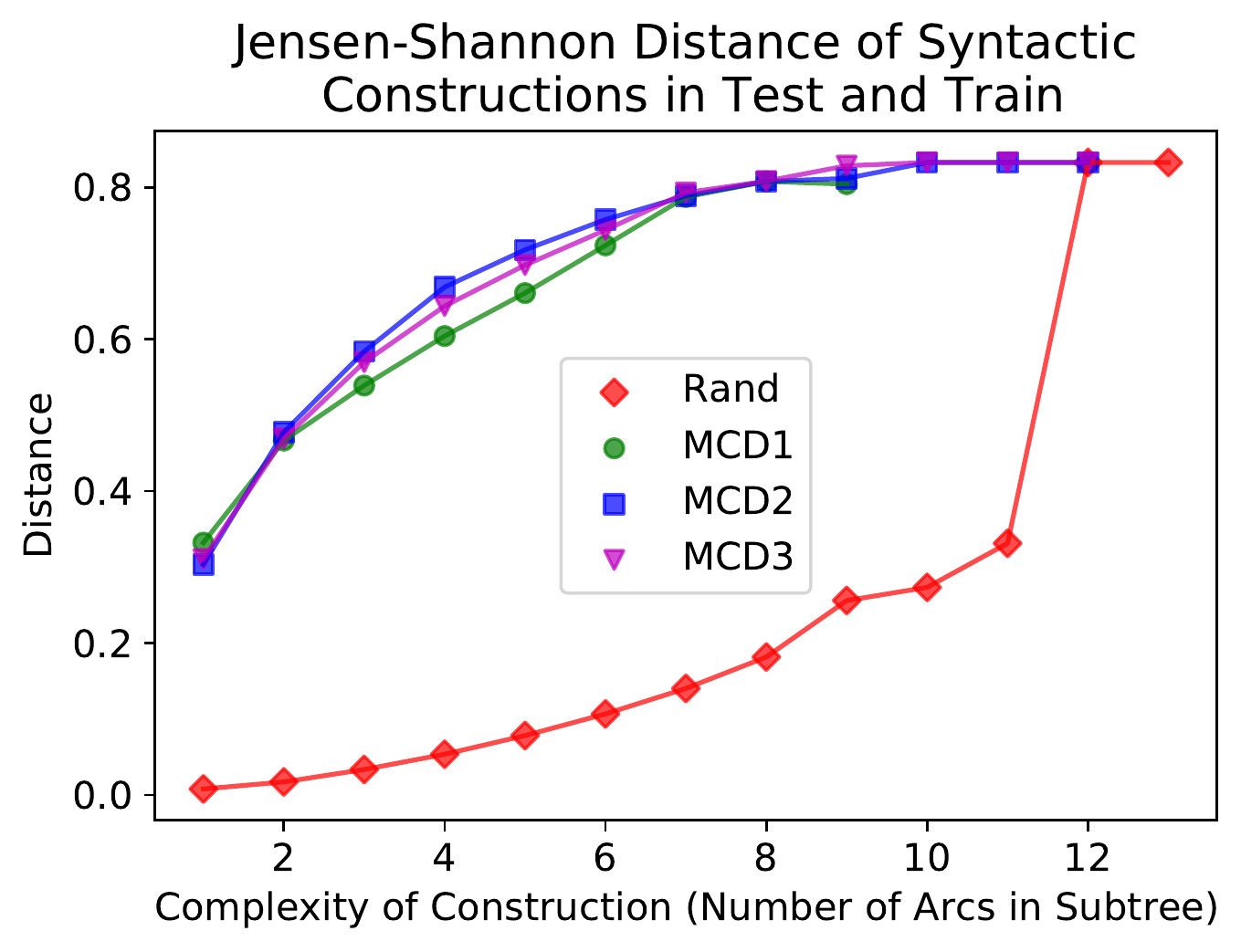}
    
    \caption{Divergence between test and train of the MCD and random splits. Roughly, divergence increases with subtree complexity, although much more rapidly for MCD splits than random. Additionally, there is little difference between the different MCD splits.}
    \label{fig:divergenceLengths}
\end{figure} The JS distances for constructions
of each complexity are plotted in
Figure~\ref{fig:divergenceLengths}. As can be seen in the figure, the
distances between test and train are similar for all MCD splits at
all subtree complexities. Even the MCD1 distances pattern with
the other MCD splits, despite the parser performance on MCD1 being
more similar to the random split. Thus, differences between the test and train
distributions at different complexities cannot explain the MCD splits' 
differential performance.

\begin{figure}[ht]
    \includegraphics[width = .45\textwidth, keepaspectratio]{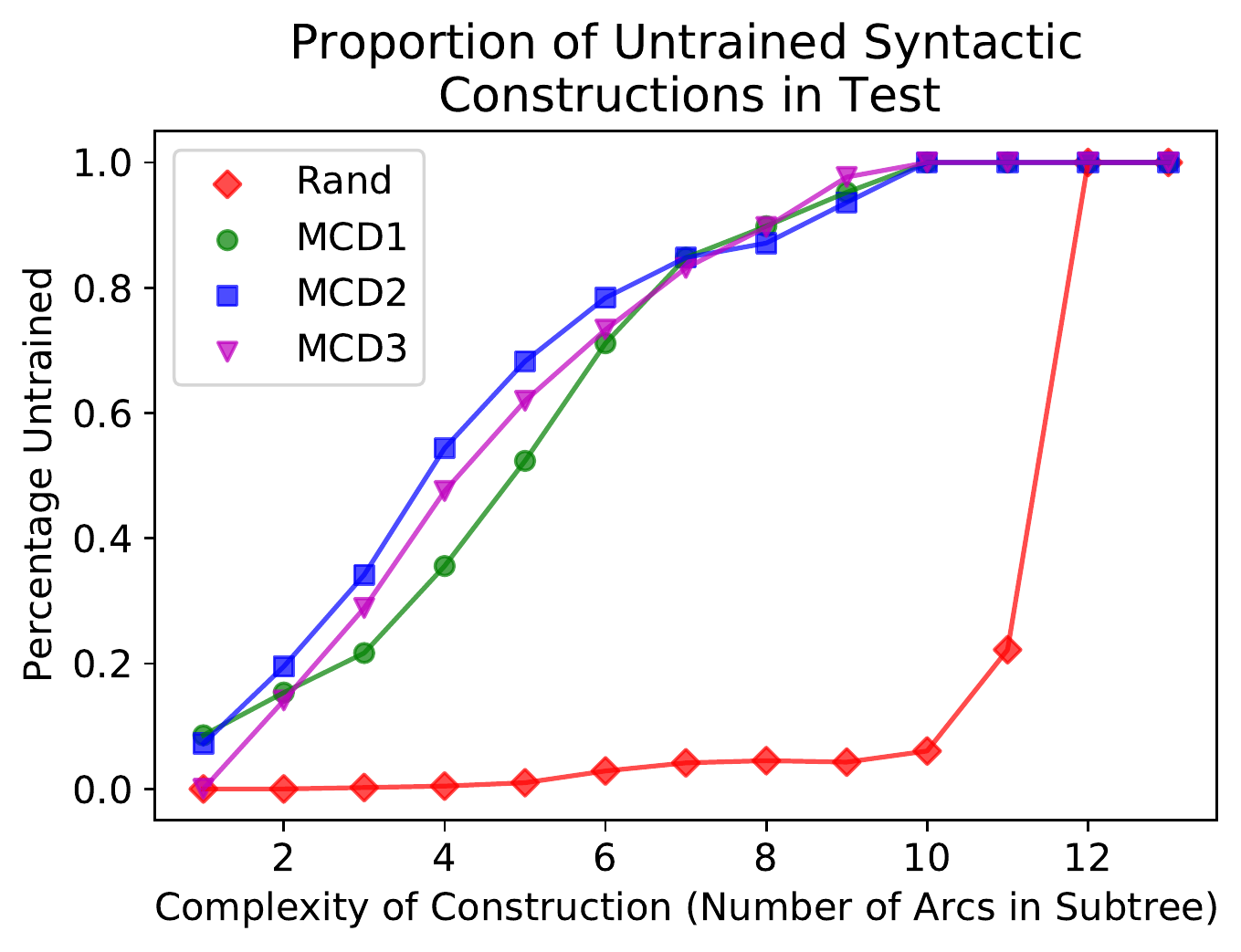}
    \caption{The proportion of syntactic constructions in test which are untrained, for all splits and all subtree complexities. Subtree complexity is measured in number of arcs.}
    \label{fig:untrainedLengths}
\end{figure}

In our second analysis, we examined whether the MCD
splits differ in the proportion of untrained subtrees at different
complexities. The proportions are plotted in
Figure~\ref{fig:untrainedLengths}. The MCD splits pattern together,
with far more untrained constructions at each complexity than the
random split.

We thus conclude it is unlikely that gross distributional
properties of the MCD splits explain the differences in parser
performance. In the next section, we show that parser
mistakes for all splits seem to be driven by a very small number
of hard-to-parse subtrees. Thus, performance differences between splits
likely depend on idiosyncratic interactions between the specific data splits and models. 

\section{Syntactic Analyses of Dependency Parser Outputs}
\label{sec:badTrees}
\subsection{Identifying Difficult Subtrees}
To identify syntactic constructions that are predictive of
dependency parsing error, we fit a logistic model predicting
Stanza's performance on each test question from the question's
syntactic constructions. Because we trained five randomly-initialized
versions of Stanza, the model was fit with five instances of each
question. To encourage sparse subtree feature weights, we used L1
regularization. We used $90\%$ of the test set to train the logistic
model, and the remaining $10\%$ to test it and select a regularization
coefficient of $.01$.

To analyze the subtrees most predictive of parsing failure, we extracted 
from the model all subtrees with a coefficient less than or equal to $-1$. 
Finally, to quantify the effect these trees have on test performance, we
removed all the sentences containing the trees for each split, and
calculated Stanza's accuracy on the remaining test sentences.

\subsection{Subtrees Predictive of Parsing Error}
\begin{table*}[htb]
\centering
\resizebox{0.8\textwidth}{!}{
\begin{tabular}{ccccccc}
\toprule
 Split & Num Trees & Sentences in & WSCLAS, StDv         & CLAS, StDv           & LAS, StDv            & Num Sentences \\
       & removed & reduced test &                      &                      &                      & removed      \\
\midrule
 mcd1  & 3 & 11,139        & 98.41 $\pm$0.71  & 99.75  $\pm$0.15 & 99.84$\pm$0.1 & 828           \\
\rowcolor{gray!6} mcd2 & 5 & 8,440         & 92.46 $\pm$3.82 & 98.59$\pm$0.47 & 98.89$\pm$0.37 & 3527                     \\
mcd3 & 7 & 5,692         & 93.09$\pm$2.77  & 98.79$\pm$0.50    & 98.91$\pm$0.36 & 6275        \\
\rowcolor{gray!6} rand  & 0 & 11,967        & 99.89$\pm$0.01  & 99.98  $\pm$0.00  & 99.99 $\pm$0.00 & 0           \\
\bottomrule
\end{tabular}}
\caption{Re-calculated accuracy on the test sets, when removing all sentences whose subtrees are most predictive of Stanza's failure.\label{tbl:reducedTestSetScores}}
\end{table*}

Table~\ref{tbl:reducedTestSetScores} shows the number of subtrees
found to be predictive of parsing error, together with the accuracy
when those trees are removed from test. Removing five subtrees from
MCD2's test set improves the accuracy to $92.46\%$ (an increase of
$21.05\%$), and removing seven trees from MCD3's test set improves the accuracy
to $93.09\%$ (an increase of $36.33\%$). We thus conclude
that the performance degradation of Stanza on higher compound
divergence splits is driven by a relatively small number of syntactic
constructions.
\begin{table*}[h!t]
\vspace{.5cm}
\small
\centering
\begin{tabular}{lllllllllll}
\toprule
 Tree &  MCD1 &       &       &    MCD2 &       &       &  MCD3 &       &    &     \\
    &  Train \# & Test \# & $\Delta \%$  & Train \# & Test \# & $\Delta \%$ & Train \# & Test \# & $\Delta \%$ &   \\
\midrule
& \multicolumn{9}{c}{Predictive for MCD1} \\
\resizebox{!}{.5in}{
\begin{dependency}
\begin{deptext}
and \& entity \& verb \\ 
\end{deptext} 
\depedge{3}{1}{{\small CC}}
\depedge{3}{2}{{\small NSUBJ}}
\deproot[edge unit distance=2ex]{3}{root}
\end{dependency}}& 2342 & 192   & 0.26  & 2099 & 144   & -0.15 & 1546 & 118   & 0.01  \\ 
 
 \resizebox{!}{.5in}{
\begin{dependency} 
\begin{deptext}
of \& entity \& entity \& entity \& entity \\
\end{deptext}

\depedge{2}{1}{{\small CASE}}
\depedge{2}{3}{{\small CONJ}}
\depedge{2}{4}{{\small CONJ}}
\depedge{2}{5}{{\small CONJ}}
\deproot[edge unit distance=2ex]{2}{root}
\end{dependency}}& 0    & 484   & 1.29  & 0    & 405   & -0.15 &  0    & 382   & 1.78  \\ 

\resizebox{!}{.5in}{
 \begin{dependency} 
\begin{deptext}
adjective \& noun \&  's \\
\end{deptext}

\depedge{2}{1}{{\small AMOD}}
\depedge{2}{3}{{\small CASE}}
\deproot[edge unit distance=2ex]{2}{root}

\end{dependency} } & 2319 & 137   & 0.23 & 3033 & 3     & -0.0  & 3255 & 50    & -0.1  \\ 

\midrule
& \multicolumn{9}{c}{Predictive for MCD2 and MCD3} \\

\resizebox{!}{.5in}{
\begin{dependency} 
\begin{deptext}
was \& role \& role \\
\end{deptext}

\depedge{3}{1}{{\small COP}}
\depedge{3}{2}{{\small NSUBJ}}
\deproot[edge unit distance=2ex]{3}{root}
\end{dependency}} & 367  & 792   & -0.0  & 0    & 940   & 5.73  &  0    & 1437  & 6.98   \\ 

\resizebox{!}{.5in}{
\begin{dependency} 
\begin{deptext}
was \& noun \& role \\
\end{deptext}

\depedge{3}{1}{{\small COP}}
\depedge{3}{2}{{\small NSUBJ}}
\deproot[edge unit distance=2ex]{3}{root}
\end{dependency}} & 0    & 530   & -0.06  & 0    & 535   & 2.25  & 0    & 593   & 0.54   \\ 

\resizebox{!}{.5in}{
\begin{dependency} 
\begin{deptext}
role \& a \& noun \\
\end{deptext}

\depedge{3}{1}{{\small NSUBJ}}
\depedge{3}{2}{{\small DET}}
\deproot[edge unit distance=1.5ex]{3}{root}

\end{dependency}} & 0    & 541   & 0.02   & 0    & 547   & 2.38  & 0    & 604   & 2.01 \\ 

\resizebox{!}{.5in}{
\begin{dependency} 
\begin{deptext}
was \& noun \& verb \\
\end{deptext}
\depedge{3}{1}{{\small AUX}}
\depedge{3}{2}{{\small NSUBJ}}
\deproot[edge unit distance=2ex]{3}{root}

\end{dependency}} & 872  & 60  & -0.02  & 0    & 512   & 2.94  &  0    & 547   & 2.61 \\ 

\resizebox{!}{.5in}{
\begin{dependency} 
\begin{deptext}
was \& role \& verb \\
\end{deptext}
\depedge{3}{1}{{\small AUX}}
\depedge{3}{2}{{\small NSUBJ}}
\deproot[edge unit distance=1.75ex]{3}{root}
\end{dependency}} & 1484 & 231   & -0.03 & 0    & 993   & 2.53  &  0    & 1093  & 3.06 \\

\midrule
& \multicolumn{9}{c}{Predictive for MCD3} \\

\resizebox{!}{.5in}{
\begin{dependency} 
\begin{deptext}
verb \& verb \& entity \& ? \\
\end{deptext}

\depedge{1}{4}{{\small PUNCT}}
\depedge{1}{3}{{\small OBJ}}
\depedge{1}{2}{{\small CONJ}}
\deproot[edge unit distance=2ex]{1}{root}

\end{dependency}} & 877  & 1513  & 0.1  & 451  & 1435  & -2.53 & 355  & 1407  & 1.47 \\ 

\resizebox{!}{.5in}{
\begin{dependency} 
\begin{deptext}
of \& entity \& entity \& entity \\
\end{deptext}

\depedge{2}{1}{{\small CASE}}
\depedge{2}{4}{{\small CONJ}}
\depedge{2}{3}{{\small CONJ}}
\deproot[edge unit distance=1.75ex]{2}{root}

\end{dependency}} & 0    & 1153  & 1.33 & 509  & 905   & -0.97 & 0    & 931   & 3.95 \\ 
\bottomrule
\end{tabular}
\caption{Syntactic constructions most predictive of dependency parsing failure for each split. ``Predictive'' means the subtree is associated with a coefficient $\leq -1$ by the logistic model. \# stands for the number of occurrences. $\Delta \%$ is defined as the change in mean WSCLAS after all instances of the construction have been removed from test. \label{tbl:badTrees}}
\vspace{.3cm}
\end{table*}

Table~\ref{tbl:badTrees} shows the subtrees most predictive of a
dependency parsing error, with their test and train frequency.
To quantify the effect of each subtree on the test accuracy, we also
report the \fm{Test set $\Delta$}:
$\textrm{WSCLAS}(T')- \textrm{WSCLAS}(T)$ where $T$ is the original
test set and $T'$ is all test sentences which do not include the
construction. A positive $\Delta$ means that removing the subtree from
the test set improved performance, while a negative $\Delta$ indicates
that removing the subtree from the test set degraded performance.

Subtrees that are predictive of error for a particular split are often missing from train, 
together with others that share a similar syntactic structure. For instance, there are a set of trees that form questions with common nouns as subject
and predicate, and a copula verb \ex{was} appearing to the left of the subject (e.g. \ex{Was an art director of Palm County a person?}).\footnote{Note that CFQ has two part-of-speech categories which are common nouns: 
a \fm{role}, like the word \ex{mother} in the phrase \ex{Henry's mother}, 
and a category labeled \fm{noun}, like \ex{person} in the phrase \ex{a person}.}
The fourth, fifth and sixth subtrees in Table~\ref{tbl:badTrees} are subtrees which form 
these questions; all three are missing from train for both MCD2 and MCD3, and all are predictive of
parser error for these splits. 
In contrast, the MCD1 training set includes one of the subtrees (fourth in Table~
\ref{tbl:badTrees}) and leaves the other two untrained; none are predictive of parser
error (with $\Delta$ of 0.0, -0.06 and 0.02, the performance on these trees is close to average for MCD1). 
The model performs better on the untrained trees in MCD1, perhaps because of the similar trees in train; 
with no evidence of this kind of structure in MCD2 and MCD3, the model struggles.\footnote{Not shown
in Table~\ref{tbl:badTrees} is the tree with a left-edge copula and simple
nouns in both predicate and subject position. This structure was also absent from train in the MCD2 and 
MCD3 splits, but present in MCD1 train. It was not found to be strongly 
predictive of errors by the logistic model, likely because it was infrequent in test (occurring 188 times in 
MCD2 and 208 in MCD3).}

Another group of subtrees with similar syntactic structure is the second and last subtrees in Table~\ref{tbl:badTrees}. These coordinate three and four entities in an \ex{of}-type prepositional phrase, which occurs in phrases like \ex{the mother of Alice, Bob, Carl and Dave}. 
Both trees are absent from MCD1 train, and both have a large effect on performance for MCD1 ($\Delta$ of 1.29 and 1.33). 
In MCD2, only the tree
with four coordinated entities is absent from train, and it is not difficult for the model ($\Delta$ of -0.15, indicating that removing it from test reduces performance);
the model is likely
able to parse four coordinated entities based on the training examples with three coordinated entities.

\section{Related work}
\label{sec:relatedwork}
A growing body of work uses CFQ to investigate better models for
compositional generalization in semantic parsing
\citep[][]{herzig2020span, guo2020hierarchical,furrer2020compositional}. 
\citet{tsarkov2020star} also recently released an expanded version of CFQ called *-CFQ, which
remains challenging for transformers even when they are trained on
much more data. Our methodology can be easily be applied to *-CFQ at
the cost of a straight-forward extension of the grammar.

Other datasets focused on compositional generalization include SCAN
\citep{lake2018generalization}, a dataset of English commands and
navigation sequences; gSCAN \citep{ruis2020benchmark}, a successor to
SCAN with grounded navigation sequences; and COGS \citep{kim2020cogs},
where English sentences are paired with semantic representations based
on lambda calculus and the UDepLambda framework
\citep{reddy2017universal}. In contrast to CFQ, these datasets
challenge models by targeting specific, linguistically-motivated
generalizations. For example, COGS includes tests of novel verb
argument structures (like training on a verb in active voice and
testing in passive voice), and novel grammatical roles for primitives
(like training with a noun in object position and testing in subject
position); similarly, SCAN includes splits which test novel
combinations of specific predicates (training a predicate \ex{jump} or
\ex{turn left} in isolation, and testing it composed with additional
predicates from train). Finally, the CLOSURE benchmark for visual
question answering tests systematic generalization of familiar words
by constructing novel referring expressions; for example, \ex{a cube
  that is the same size as the brown
  cube} \citep{bahdanau2020closure}.

\section{Conclusion}
In this paper, we presented a dependency parsing version of the
Compositional Freebase Queries (CFQ) dataset. We showed that a
state-of-the-art dependency parser's performance degrades with
increased compound divergence, but varies on different splits of the
same compound divergence. Finally, we showed the majority of the parser
failures on each split can be characterized by a small (seven or
fewer) number of specific syntactic structures.

To our knowledge, this is the first explicit test of compositional
generalization in dependency parsing. We hope that the gold-standard
dependency parses that we have developed will be a useful resource in
future work on compositional generalization. Existing work on
syntactic (and in particular dependency) parsing can provide
researchers in compositional generalization with ideas and inspiration
which can then be empirically validated using our corpus.

Finally, our work represents a step forward in understanding the
syntactic structures which drive lower performance on MCD test
sets. Predicting parser performance from the syntactic
constructions contained in the question provides a new method for
understanding the syntactic structures that can cause parser failure; in
future work, similar methods can also be used to better understand
failures of semantic parsers on the CFQ dataset.

\section*{Ethical Considerations}
This article contributes to compositional generalization research, a foundational concern for neural natural natural language processing models. Breakthroughs in this research might eventually lead to smaller, and more efficient models, as well as better performance on low-resource languages. The ethical and societal consequences of these improvements will depend on downstream applications.

The resource released in this work is a new set of annotations for CFQ, an existing dataset. The original CFQ dataset was artificially generated, so there was no process of data collection and therefore no ethics review process. The dataset was annotated by the authors, so there was no ethics review of the annotation process or demographic information of this population to report. 

\section*{Acknowledgements}
We thank Christopher Manning, the Montreal Computational and Quantitative Linguistics lab at McGill University, and the Human and Machine Interaction Through Language group at ServiceNow Research for helpful feedback.
We are grateful to ServiceNow Research for providing extensive compute and other support. We also gratefully acknowledge the support of the MITACS accelerate internship, the Natural Sciences and Engineering Research Council of Canada, 
the Fonds de Recherche du Qu\'{e}bec, Soci\'{e}t\'{e} et Culture, 
and the Canada CIFAR AI Chairs Program. 

\bibliography{refs}
\bibliographystyle{acl_natbib}

\appendix

\section{Correlation of Semantic Parsing and Dependency Parsing Errors}
Because syntactic parsing is a necessary sub-task for semantic parsing, we also explored the possibility that dependency parsing errors might be predictive of semantic parsing errors. We extracted the predictions from \citet{keysers2020measuring}'s transformer model (which is based on \citet{vaswani2017attention}'s model), and compared them to those of Stanza on the same test set. For each test sentence, we calculated the number of times the parsers correctly parsed the sentence (out of five experiments each). 

\begin{figure}[h]
\includegraphics[width=.23\textwidth]{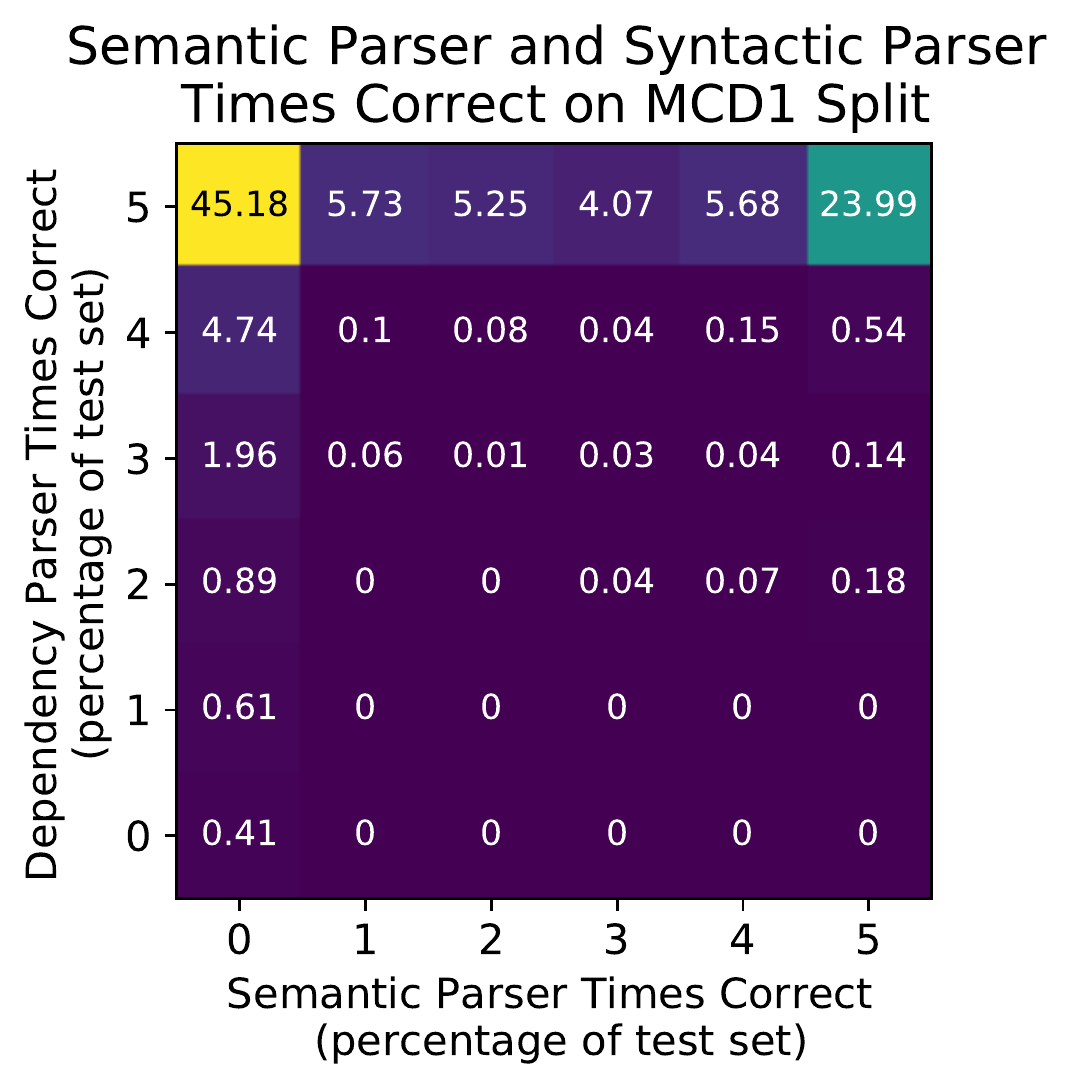}
\includegraphics[width=.23\textwidth]{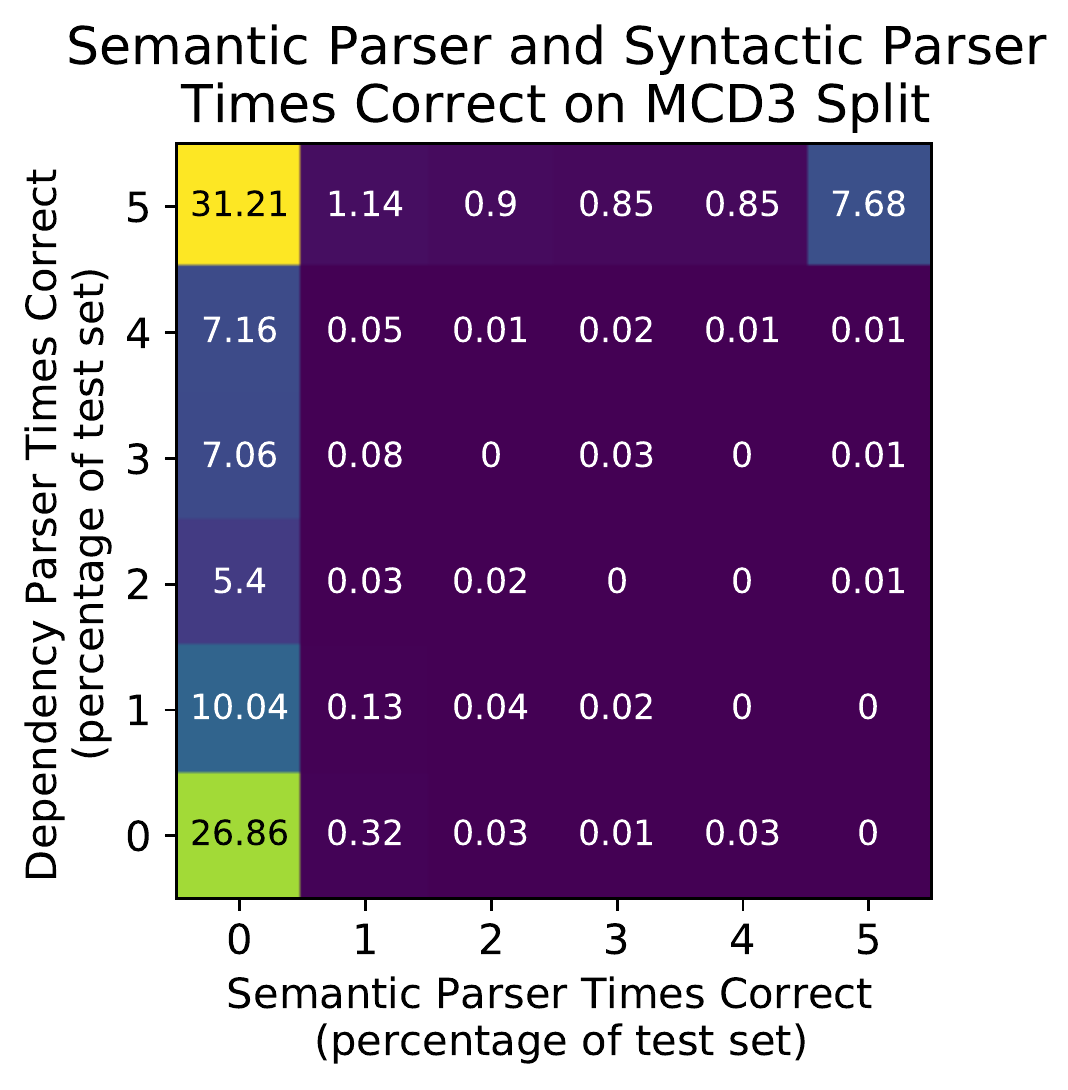}
\caption{Percentage of the test set which is correctly parsed by five semantic parsing experiments and five dependency parsing experiments, for MCD1 and MCD3.}
\label{fig:errormatrices}
\end{figure}
The results for MCD1 and MCD3 are shown in Figure~\ref{fig:errormatrices}: for example, the top-right hand corner of the MCD1 matrix means that 23.99\% of the test set was correctly parsed in all semantic and dependency parsing experiments, while the top right-hand corner in the MCD3 matrix indicates that only 7.68\% of the sentences were correctly parsed by both models in all experiments. The semantic parser fails for all five experiments on the majority of sentences. We do note some trends in error patterns between the models: for example, no sentences are correctly parsed by all semantic parsers without also being correctly parsed by the dependency parser at least a few times. However, overall it does not appear that dependency parsing performance is strongly related to  semantic parsing performance.

\section{Proportion of Few-shot Constructions in MCD Splits}
We examined whether the MCD splits differ in the proportion of test syntactic constructions which are \fm{few-shot}, meaning they appear in train fewer than four times. This analyses is similar to the ones described in \textsection~\ref{sec:split_tree_distributions}. 

\begin{figure}[ht]
    \includegraphics[width = .45\textwidth,keepaspectratio]{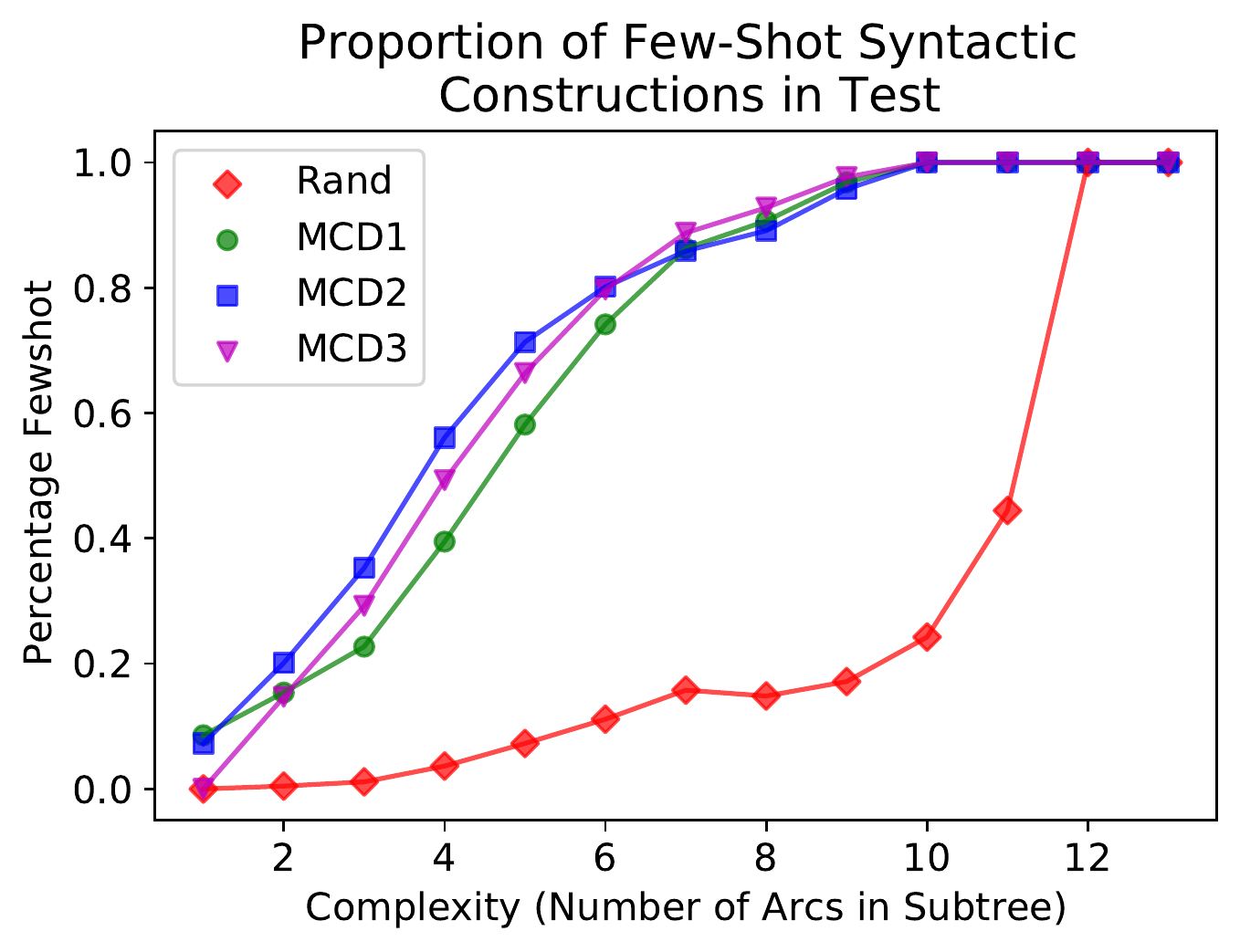}
    \caption{The proportion of syntactic constructions in test which appear in train fewer than four times, for all subtree complexities. Complexity is measured in number of arcs.}
    \label{fig:fewshotLengths}
\end{figure}

The proportions are plotted in Figure~\ref{fig:fewshotLengths}. The MCD splits pattern together, with far more few-shot constructions at each complexity than the random split.

\section{CFG}
\label{sec:CFG}
Below are the rules in our Context Free Grammar. Using these rules we parsed CFQ into constituency trees, and then mapped to dependency trees as described in \textsection~\ref{subsec:goldparses}.

S $\longrightarrow$ NPQ VP Qmark\newline
S $\longrightarrow$ NPQ was Nominal Qmark\newline
S $\longrightarrow$ NPQ did NPV Qmark\newline
S $\longrightarrow$ was Nominal Vobl Qmark\newline
S $\longrightarrow$ NPQ Vobl Qmark\newline
S $\longrightarrow$ was Nominal Adj Qmark\newline
S $\longrightarrow$ was Nominal Nominal Qmark\newline
S $\longrightarrow$ did Nominal VP Qmark\newline \newline
NPV $\longrightarrow$ Nominal V \newline
NPV $\longrightarrow$ Nominal VPrep\newline \newline
VP $\longrightarrow$ V Nominal\newline
VP $\longrightarrow$ was Vobl\newline
VPrep $\longrightarrow$ was VPrep\newline
VPrep $\longrightarrow$ V by\newline
Vobl $\longrightarrow$ VPrep Nominal \newline
NPQ $\longrightarrow$ WhW Nominal\newline
NPQ $\longrightarrow$ WhW role caseO\newline \newline
commonNoun $\longrightarrow$ commonNoun RC\newline \newline
RC $\longrightarrow$ Vobl\newline
RC $\longrightarrow$ R VP\newline
RC $\longrightarrow$ R NPV\newline
RC $\longrightarrow$ whose role VP\newline \newline
VP $\longrightarrow$ VP andVP \newline
VP $\longrightarrow$ VPx andVP \newline
VPx $\longrightarrow$ VP punctVP\newline
VPx $\longrightarrow$ VPx punctVP\newline
andVP $\longrightarrow$ conj VP \newline
andVP $\longrightarrow$ punct conj VP\newline
punctVP $\longrightarrow$ punct VP\newline \newline
Vobl $\longrightarrow$ Vobl andVobl \newline
Vobl $\longrightarrow$ Voblx andVobl\newline
Voblx $\longrightarrow$ Vobl punctVobl\newline
Voblx $\longrightarrow$ Voblx punctVobl\newline
andVobl $\longrightarrow$ conj Vobl\newline
andVobl $\longrightarrow$ punct conj Vobl\newline
punctVobl $\longrightarrow$ punct Vobl\newline \newline
VPrep $\longrightarrow$ VPrep andVPrep\newline
VPrep $\longrightarrow$ VPrepX andVPrep\newline
VPrepX $\longrightarrow$ VPrep punctVPrep \newline
VPrepX $\longrightarrow$ VPrepX punctVPrep\newline
andVPrep $\longrightarrow$ conj VPrep \newline
andVPrep $\longrightarrow$ punct conj VPrep\newline
punctVPrep $\longrightarrow$ punct VPrep\newline \newline
V $\longrightarrow$ V andV\newline
V $\longrightarrow$ Vx andV\newline
Vx $\longrightarrow$ V punctV\newline
Vx $\longrightarrow$ Vx punctV\newline
andV $\longrightarrow$ conj V\newline
andV $\longrightarrow$ punct conj V\newline
punctV $\longrightarrow$ punct V\newline \newline
Vx $\longrightarrow$ Vx punctVPrep\newline
Vx $\longrightarrow$ V punctVPrep\newline
V $\longrightarrow$ Vx andVPrep\newline
V $\longrightarrow$ V andVPrep\newline
VPrep $\longrightarrow$ VPrep andV\newline
VPrep $\longrightarrow$ VPrepX andV\newline
VPrepX $\longrightarrow$ VPrep punctV\newline
VPrepX $\longrightarrow$ VPrepX punctV\newline \newline
NPV $\longrightarrow$ NPV andNPV\newline
NPV $\longrightarrow$ NPVx andNPV\newline
NPVx $\longrightarrow$ NPV punctNPV\newline
NPVx $\longrightarrow$ NPVx punctNPV\newline
andNPV $\longrightarrow$ conj NPV\newline
andNPV $\longrightarrow$ punct conj NPV \newline
punctNPV $\longrightarrow$ punct NPV\newline \newline
V $\longrightarrow$ F V\newline \newline
Nominal $\longrightarrow$ Name\newline
Nominal $\longrightarrow$ DP\newline
Nominal $\longrightarrow$ commonNoun\newline \newline
DP $\longrightarrow$ caseS role\newline
caseS $\longrightarrow$ DP pS \newline
caseS $\longrightarrow$ Name pS \newline
DP $\longrightarrow$ det role caseO\newline
caseO $\longrightarrow$ of DP\newline
caseO $\longrightarrow$ of Name\newline \newline
DP $\longrightarrow$ det commonNoun\newline \newline \newline
Name $\longrightarrow$ Name andName\newline
Name $\longrightarrow$ Namex andName\newline
Namex $\longrightarrow$ Name punctName\newline
Namex $\longrightarrow$ Namex punctName\newline
andName $\longrightarrow$ conj Name\newline
andName $\longrightarrow$ punct conj Name\newline
punctName $\longrightarrow$ punct Name\newline \newline
commonNoun $\longrightarrow$ commonNoun andCommonNoun\newline
commonNoun $\longrightarrow$ commonNounx andCommonNoun\newline
commonNounx $\longrightarrow$ commonNoun punctCommonNoun\newline
commonNounx $\longrightarrow$ commonNounx punctCommonNoun\newline
andCommonNoun $\longrightarrow$ conj commonNoun \newline
andCommonNoun $\longrightarrow$ punct conj commonNoun\newline
punctCommonNoun $\longrightarrow$ punct commonNoun \newline \newline
role $\longrightarrow$ role androle\newline
role $\longrightarrow$ rolex androle\newline
rolex $\longrightarrow$ role punctrole\newline
rolex $\longrightarrow$ rolex punctrole\newline
androle $\longrightarrow$ conj role \newline
androle $\longrightarrow$ punct conj role\newline
punctrole $\longrightarrow$ punct role\newline \newline
commonNoun $\longrightarrow$ F commonNoun\newline
role $\longrightarrow$ F role\newline
role $\longrightarrow$ Cnt of nat\newline
commonNoun $\longrightarrow$ P commonNoun\newline \newline
commonNoun $\longrightarrow$ Adj commonNoun\newline
role $\longrightarrow$ Adj role\newline \newline \newline
punct $\longrightarrow$ ,\newline
Cnt $\longrightarrow$ country\newline
nat $\longrightarrow$ nationality \newline
P $\longrightarrow$ production\newline
F $\longrightarrow$ film | art | executive | costume \newline
V $\longrightarrow$ VP$\_$SIMPLE | direct | produce ...\newline
Name $\longrightarrow$ entity  | Alice | Bob ... \newline
commonNoun $\longrightarrow$ NP$\_$SIMPLE | character | person ... \newline
role $\longrightarrow$ ROLE$\_$SIMPLE | character | person ... \newline
NPQ $\longrightarrow$ who | what\newline
WhW  $\longrightarrow$ What | Which | what | which\newline
did $\longrightarrow$ did | Did\newline
conj $\longrightarrow$ and\newline
pS $\longrightarrow$ `s\newline
of $\longrightarrow$ of\newline
det $\longrightarrow$ a | an\newline
by $\longrightarrow$ by\newline
Adj $\longrightarrow$ ADJECTIVE$\_$SIMPLE | female | American ... \newline
was $\longrightarrow$ was | were \newline
R $\longrightarrow$ that \newline
whose $\longrightarrow$ whose\newline
Qmark $\longrightarrow$ ?
\end{document}